\documentclass{article}
\usepackage{spconf,amsmath,graphicx}
\usepackage{enumitem}
\usepackage{amssymb}
\usepackage[font=small,labelfont=bf, justification=justified,
format=plain]{caption} 
\usepackage{cite}
\usepackage{fancyhdr}

\title{Person Re-identification using Visual Attention}
\name{Alireza Rahimpour, Liu Liu, Ali Taalimi, Yang Song,  Hairong Qi}
\address{Department of Electrical Engineering and Computer Science\\
	University of Tennessee, Knoxville, TN, USA 37996\\
	$ \{ $arahimpo, lliu25, ataalimi, ysong18, hqi$ \} $@utk.edu}
\usepackage{lipsum}
\usepackage{fancyhdr}
\begin{document}

%
\maketitle

\begin{abstract}
Despite recent attempts for solving the person re-identification problem, it remains a challenging task since a person's appearance can vary significantly when large variations in view angle, human pose and illumination are involved. In this paper, we propose a novel approach based on using a gradient-based attention mechanism in deep convolution neural network for solving the person re-identification problem. Our model learns to focus selectively on parts of the input image for which the networks' output is most sensitive to and processes them with high resolution while perceiving the surrounding image in low resolution.
Extensive comparative evaluations demonstrate that the proposed method outperforms state-of-the-art approaches on the challenging CUHK01, CUHK03 and Market 1501 datasets.
\end{abstract}
\section{Introduction}
\label{sec:intro}
Recently, person re-identification has gained increasing research interest in the computer vision community due to its importance in multi-camera surveillance systems. Person re-identification is the task of matching people across non-overlapping camera views at different times.
A typical re-identification system takes as input two images of person's full body, and outputs either a similarity score between the two images or the decision of whether the two images belong to the same identity or not. 
Person re-identification is a challenging task since different individuals can share similar appearances and also appearance of the same person can be drastically different in two different views due to several factors such as background clutter, illumination variation and pose changes.

It has been proven that humans do not focus their attention on an entire scene at once when they want to identify another person \cite{xu2015show}. Instead, they \textit{pay attention} to different parts of the scene (e.g., the person's face) to extract the most discriminative information. Inspired by this observation, we study the impact of attention mechanism in solving person re-identification problem. The attention mechanism can significantly reduce the complexity of the person re-identification task, where the network learns to focus on the most informative regions of the scene and ignores the irrelevant parts such as background clutter. Exploiting the attention mechanism in person re-identification task is also beneficial at scaling up the system to large high quality input images.  

With the recent surge of interest in deep neural networks, attention based models have been shown to achieve promising results on several challenging tasks, including caption generation \cite{xu2015show}, machine translation \cite{bahdanau2014neural} and object recognition \cite{ba2014multiple}. 
However, attention models proposed so far, require defining an explicit predictive model, whose training can pose challenges due to the non-differentiable cost. Furthermore, these models employ Recurrent Neural Network (RNN) for the attention network and are computationally expensive or need some specific policy algorithms such as REINFORCE \cite{ba2014multiple, williams1992simple} for training. 

In this paper, we introduce a novel model architecture for person re-identification task which improves the matching accuracy by taking advantage of attention mechanism. 
The contributions of this research are the following:
\begin{itemize}
	\item {We propose a CNN-based attention model which is specifically tailored for the person re-identification task in a triplet loss architecture. Furthermore, our deep model is interpretable thanks to the generated attention maps.} 
	\item The model is easy to train and is computationally efficient during inference, since it first finds the most discriminative regions in the input image and then performs the high resolution feature extraction only on the selected regions.  
	\item Finally, we qualitatively and quantitatively validate the performance of our proposed model by comparing it to the
	state-of-the-art performance on three challenging benchmark datasets: CUHK01\cite{CH01li2013locally}, CUHK03 \cite{li2014deepreid} and Market 1501 \cite{zheng2015scalable}.
\end{itemize}

\section{Related works}
Generally, existing approaches for person re-identification are mainly focused on two aspects: learning a distance metric \cite{ liao2015person,pedagadi2013local, su2015multi} and developing a new feature representation \cite{varior2016learning, zhao2013unsupervised, liao2010modeling, ojala2002multiresolution, zhao2013person, zheng2015scalable}. In distance metric learning methods, the goal is to learn a metric that emphasizes inter-personal distance and de-emphasizes intra-person distance. The learnt metric is used to make the final decision as to whether a person has been re-identified or not (e.g.,
KISSME \cite{kostinger2012large}, XQDA \cite{liao2015person}, MLAPG \cite{su2015multi} and LFDA \cite{pedagadi2013local}).
In the second group of methods based on developing new feature representation for person re-identification, novel feature representations were proposed to address the challenges such as variations in illumination, pose and view-point \cite{varior2016learning}. The Scale Invariant Local Ternary Patterns (SILTP) \cite{liao2010modeling}, Local Binary Patterns (LBP) \cite{ojala2002multiresolution}, Color Histograms \cite{zhao2013person} or Color Names \cite{zheng2015scalable} (and combination of them), are the basis of the majority of these feature representations developed for human re-identification. 

In the recent years, several approaches based on Convolutional Neural Network (CNN) architecture for human re-identification have been proposed and achieved great results \cite{cheng2016person, li2014deepreid, ahmed2015improved}. In most of the CNN-based approaches for re-identification, the goal is to jointly learn the best feature representation and a distance metric (mostly in a Siamese fashion \cite{bromley1993signature}). With the recent development of RNN networks, the attention-based models have demonstrated outstanding performance on several challenging tasks including action recognition \cite{sharma2015action}. At the time of writing this research, except for one recent work \cite{liu2016end}, the attention mechanism has not yet been studied in the person re-identification literatures. In \cite{liu2016end}, the RNN-based attention mechanism is based on the attention model introduced in \cite{sharma2015action} for action recognition.

Different from \cite{liu2016end}, in our model the selection of the salient regions is made using a novel gradient-based attention mechanism, that efficiently identifies the input regions for which the network's output is most sensitive to. 
Moreover, our model does not use the RNN architecture as in \cite{liu2016end}, thus is computationally more efficient and easier to train. Furthermore, in \cite{liu2016end} the attention model requires a set of multiple glimpses to estimate the attention which is not required in our proposed architecture. 

\section{Model Architecture}
\label{sec:Method}
\begin{figure}[th!]
	\includegraphics[width=1\linewidth]{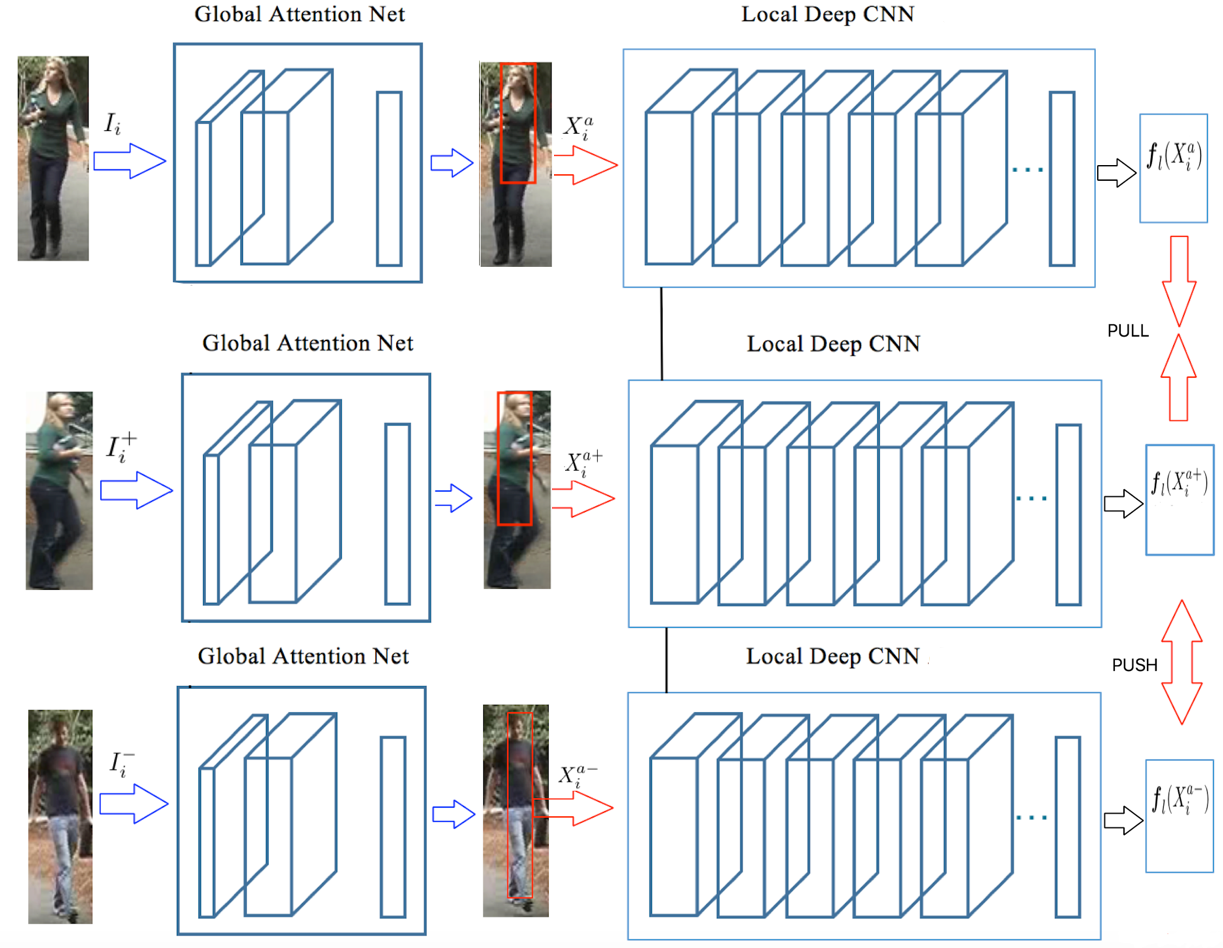}
	\caption{{The architecture of the proposed Gradient-based Attention Network in training phase. First images go through the global attention network (low resolution) and salient parts of the images are detected. Then the local deep CNN network (high resolution) is applied only over the attended regions of the input image $I_i$ determined by the attention network (i.e., $X_i^a$, shown with red boxes)}. The triplet loss forces the distance between feature representation of image pairs of the same person to be minimum and the distance of negative feature pairs to be maximum. Best viewed in color.} 
	\label{fig:fig1_0}
	
\end{figure}
In this section we introduce our gradient-based attention model within a triplet comparative platform specifically designed for person re-identification.
We first describe the overall structure of our person re-identification design, then we elaborate on the network architecture of the proposed attention mechanism. 

\subsection{Triplet Loss}
We denote the triplets of images by $<I_{i}^{+},I_{i}^{-},I_{i}>$, where $I_{i}^{+}$ and $I_{i}$ are images from the same person and $I_{i}^{-}$ is the image from a different person. As illustrated in Figure \ref{fig:fig1_0}, each image initially goes through the global attention network and salient regions of the image are selected (i.e., $X^a$). Then only these selected regions of the image pass through the local deep CNN. The local CNN network then maps this raw image regions to the feature space $<{f}_l(X_{i}^{a+}),{f}_l(X_{i}^{a-}),{f}_l(X^a_{i})>$, such that the distance of the learned features of the same person is less than the distance between the images from different persons by a defined margin. Hence, the goal of the network is to minimize the following cost function for $N$ triplet images:  
\begin{equation} \label{eq:Triplet loss}
\small
\centering
\begin{gathered}
	J=\frac{1}{N}\sum_{i=1}^{N}max(\left\|{f}_l(X^a_{i})- {f}_l(X_{i}^{a+})\right\|^2_2 - \\ \left\|{f}_l(X^a_{i})-{f}_l(X_{i}^{a-})\right\|^2_2  
	+  \alpha, 0),
\end{gathered}
\end{equation} 
where $\alpha$ is a predefined margin which helps the model to learn more discriminative features. Choosing the right triplets is critical in training of the triplet loss. For instance, if we use easy negative and positive samples for each anchor, the loss would be zero all the time and the model will not learn anything during training. 
We define the hard triplets as the triplets where the distance of the negative sample embedding to the anchor embedding is less than the distance of the positive sample embedding to the anchor embedding. We also define semi-hard triplets as triplets that satisfy the following inequality:
\begin{equation} \label{eq:Triplet loss2}
\small
\centering
\begin{gathered}
\left\|{f}_l(X^a_{i})- {f}_l(X_{i}^{a+})\right\|^2_2 < \left\|{f}_l(X^a_{i})-{f}_l(X_{i}^{a-})\right\|^2_2  <\\ \left\|{f}_l(X^a_{i})- {f}_l(X_{i}^{a+})\right\|^2_2 +  \alpha
\end{gathered}
\end{equation}
 
For training of our model we follow the hard and semi-hard negative sample mining based on the framework proposed in \cite{schroff2015facenet}.
It is important to note that the above triplet architecture is used only in the training phase and during testing, the distances between embedding of the query and gallery images are computed and used for ranking.

\subsection{Gradient-based Attention Network}
The proposed Gradient-based Attention Network (GAN) is capable of extracting information from an image by adaptively selecting the most informative image regions and only processing the selected regions at high resolution. The whole model  comprises of two blocks: the global attention network $G$ and the local deep CNN network $L$. 
The global network consists of only two layers of convolution and is computationally efficient, whereas the local network is deeper (e.g., many convolutional layers) and is computationally more expensive, but has better performance. 

We refer to the feature representation of the global layer and the local layer by ${f}_g$ and ${f}_l$, respectively. The attention model uses backpropagation to identify the few vectors in the global feature representation ${f}_g(I)$ to which the distribution over the output of the network (i.e., ${\boldsymbol{h}}_g$) is most sensitive. In other words, given the input image $I$,  ${f}_{g}(I) =\{{\boldsymbol{g}}_{i,j} | (i,j)\in [1, d_1] \times[1,d_2] \}$, where $d_1$ and $d_2$ are spatial dimensions that depend on the image size and ${\boldsymbol{g}}_{i,j} = {f}_{g}(x_{i,j}) \in \mathbb{R}^D$ is a feature vector associated with the input region $(i, j)$ in $I$, i.e., corresponds to a specific receptive field or a patch in the input image. On top of the convolution layers in attention model, there exists a fully connected layer followed by a max pooling and a softmax layer, which consider the bottom layers' representations ${f}_g(I)$ as input and output a distribution over labels, i.e., $\boldsymbol{h}_g$. 

Next, the goal is to estimate the attention map. We use the entropy of the output vector $\boldsymbol{h}_g$ as a measure of saliency in the following form: 

\begin{equation} \label{eq:entropy}
\centering
\begin{gathered}
H=\sum_{l=1}^{C}{\boldsymbol{h}_{g}^{l}\log (\boldsymbol{h}_{g}^{l})},
\end{gathered}
\end{equation} 
where $C$ is the number of class labels in the training set. In order to find the attention map we then compute the norm of the gradient of the entropy $H$ with respect to the feature vector $\boldsymbol{g}_{i,j}$ associated with the input region $(i,j)$ in the input image: 
\begin{equation} \label{eq:Gradient}
\centering
\begin{gathered}
A_{i,j}= \left\|\nabla_{\boldsymbol{g}_{i,j}}{H}\right\|_2,
\end{gathered}
\end{equation}
hence, the whole attention map would be $\boldsymbol{A} \in \mathbb{R}^{d_1 \times{d_2}}$ for the whole image. Using the attention map $\boldsymbol{A}$, we select a set of $k$ input region positions $(i,j)$ corresponding to the $A_{i,j}$s with the $k$ largest values. The selected regions of the input image corresponding to the selected positions are denoted by $X^a = \{x_{i,j}|(i,j) \in [1, d_1] \times[1,d_2]\} $, where each $x_{i,j}$ is a patch in input image $I$. 
Exploiting the gradient of the entropy as the saliency measure for our attention network encourages selecting the input regions which have the maximum effect on the uncertainty of the model predictions. 
Note that all the elements of the attention map $\boldsymbol{A}$ can be calculated efficiently using a single pass of backpropagation. For training of the global attention network ($G$), we maximize the log-likelihood of the correct labels (using cross-entropy objective function). 

After selecting the salient patches ($X^a$) within the input image, the local deep network ($L$) will be applied only on those patches. This leads to major saving in computational cost of the network and accuracy improvement by focusing on the informative regions of the person's image. The local deep CNN network ($L$) is trained on attended parts of the input image using the triplet loss introduced in Eq. \ref{eq:Triplet loss}. 
We denote the feature representation created by the local deep network $L$ as ${f}_l(X^a)$. 
 
In the test time, the local feature representation ${f}_l(X^a)$ and the global feature representation ${f}_g(I)$ are fused to create a refined representation of the whole image. 
In feature fusion, we replace the global features (low resolution features) corresponding to the attended regions (i.e., $X^a$) with the rich features from the deep CNN (high resolution features). Fusion of the features which are trained based on two discriminative losses leads to highly accurate retrieval performance. 

\section{Experiments and Results}
\subsection{Network Design}
We implement our network using TensorFlow \cite{tensorflow2015-whitepaper} deep learning framework. The training of the GAN converges in roughly $6$ hours on Intel Xeon CPU and NVIDIA TITAN X GPU. 
In the global attention network, there are $2$ convolutional layers, with $7 \times 7$ and $3 \times 3$ filter sizes, $12$ and $24$ filters, respectively. On the top of the two convolution layers in the global attention network there are one fully connected layer, a max pooling and a softmax layer. The global attention network is trained once for the whole network with cross-entropy loss. The set of
selected patches $X^a$ is composed of eight patches of size $14 \times 14$ pixels (experiments showed that the marginal improvement becomes insignificant beyond $8$ patches). The Inception-V3 \cite{szegedy2016rethinking} model pretrained on Imagenet is used for the local deep CNN.  

Inception-V3 is a $48$-layer deep convolutional architecture and since it employs global average pooling instead of fully-connected layer, it can operate on arbitrary input image sizes. The output of the last Inception block is aggregated via global average pooling to produce the feature embedding.
We use Batch Normalization \cite{ioffe2015batch} and Adam \cite{kingma2014adam} for training our model. We have employed the same scheme for data augmentation as in \cite{cheng2016person}. Furthermore, we have used $\alpha=0.02$ in Eq. \ref{eq:Triplet loss} and exponential learning rate decay for the training (initial learning rate: $0.01$).

\subsection{Datasets}

There are several benchmark datasets for evaluation of different person re-identification algorithms. In this research we use CUHK01 \cite{CH01li2013locally}, CUHK03 \cite{li2014deepreid} and Market 1501 \cite{zheng2015scalable} which are three of the largest benchmark datasets suitable for training the deep convolutional network. 

\textbf{CUHK01 dataset} contains $971$ persons captured from two camera views in a campus environment. Camera view $A$ captures frontal or back views of a person while camera $B$ captures the person's profile views. Each person has four images with two from each camera. We use $100$ persons for testing and the rest for training. 

\textbf{CUHK03 dataset} contains $13,164$ images of $1,360$ identities. All pedestrians are captured by six cameras, and each person's image is only taken from two camera views. It consists of manually cropped person images as well as images that are automatically detected for simulating more realistic experiment situation. In our experiments we used the cropped person images. We randomly select $1160$ persons for training and $100$ persons for testing.

\textbf{Market1501} dataset contains $32,688$ bounding boxes of $1,501$ identities, most of which are cropped by an automatic pedestrian detector. Each person is captured by $2$ to $6$ cameras and has $3.6$ images on average at each viewpoint. In our experiments, $750$ identities are used for training and the remaining $751$ for testing.

\subsection{Evaluation Metric and Results}
\begin{figure}[h!]
	\centering
	\includegraphics[width=1\linewidth]{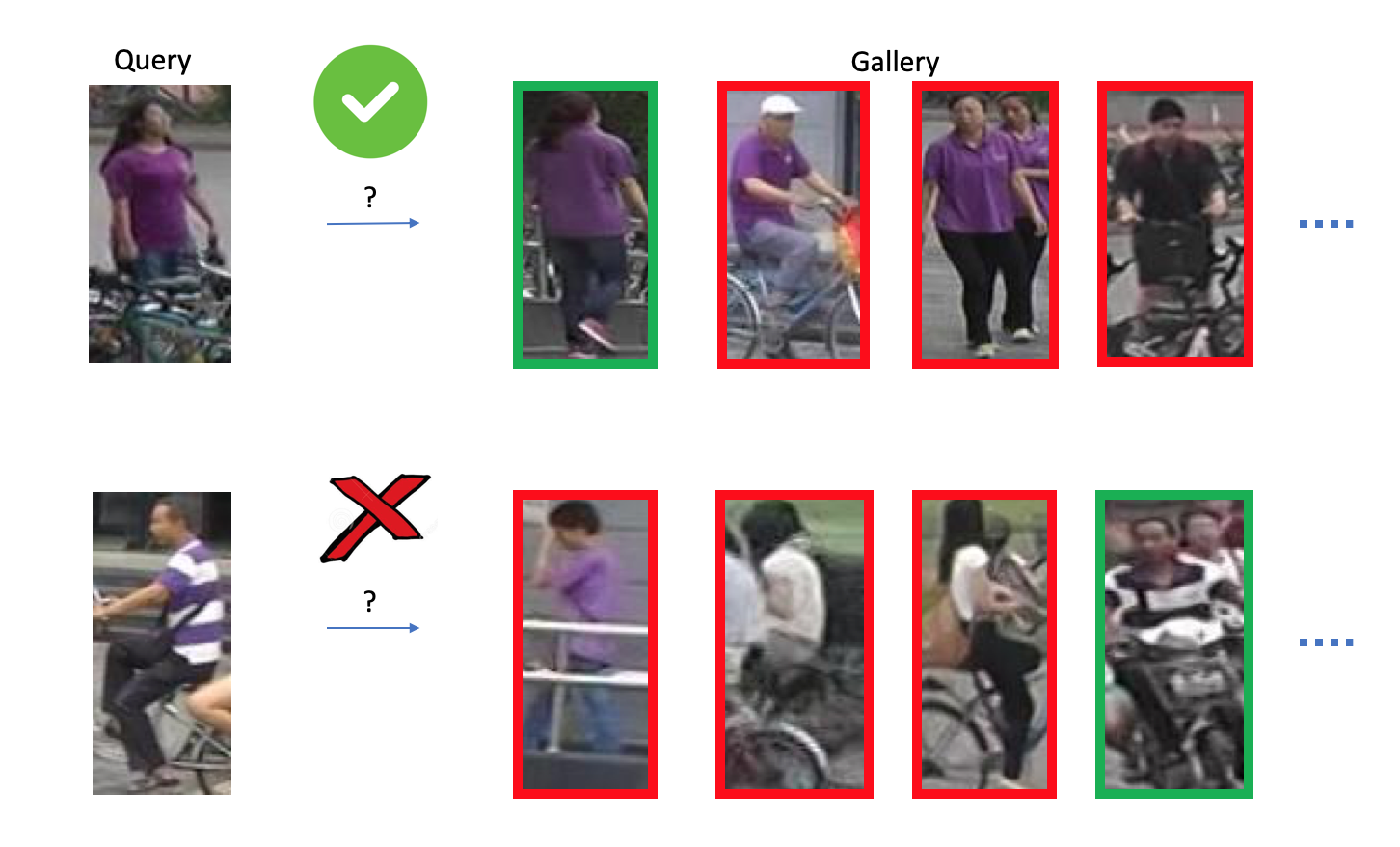}
	\caption{Rank1 retrieval on Market 1501. Top figure shows an example of successful retrieval using our model and the bottom figure shows a fail case for rank1. However in the fail case, GAN can still retrieve the image in rank4. left images are the query and the right images are the ranked images using GAN.}
	\label{fig:fig10}
\end{figure}

We adopt the widely used Rank1 accuracy for quantitative evaluations. Moreover, since the mean Average Precision (mAP) has been used for evaluation on Market 1501 data set in previous works, we use mAP for performance comparison on Market 1501 as well. 
For datasets with two cameras, we randomly select one image of a person from camera $A$ as a query image and one image of the same person from camera $B$ as a gallery image. For each image in the query set, we first compute the distance between the query image and all the gallery images using the Euclidean distance and then return the top $n$ nearest images in the gallery set. If the returned list contains an image featuring the same person as that in the query image at $k$-th position, then this query is considered as success of rank $k$.
Table \ref{table:acc} shows the rank1 accuracy of our model compared to state-of-the-art. It can be observed that the GAN (ours) outperforms all the other methods. One success (top) and fail (bottom) case in rank1 retrieval on Market 1501 data set using GAN is shown in Figure \ref{fig:fig10}.

Furthermore, GAN is computationally more efficient compared to the case where the local CNN is applied on the whole input image. In practice we observed a time speed-up by a factor of about $2.5$ by using GAN (fusion of local and global features) in test stage (tested on $100$ test images).

\begin{table}[h]
	\small
	\captionsetup{justification=centering}
	\caption{
		Rank1 accuracy (\%) comparison of the proposed method
		to the state-of-the-art. \label{table:acc}  }
	\centering
	\begin{tabular}{|c|c|c|c|c|c|}
		\hline 
		Method& Market 1501 & CUHK01 & CUHK03 & mAP\\ 
		\hline 
		KISSME \cite{kostinger2012large} & - & 29.40 &14.12&19.02\\
		GS-CNN \cite{varior2016gated}  &65.88&-&61.08&39.55\\
		DGD \cite{xiao2016learning}  &59.53&-&-&31.94\\
		LS-CNN \cite{varior2016siamese} & 61.60&-&57.30&35.30\\
		SCSP   \cite{chen2016similarity} &51.9 &- &-&26.35\\
		DNS \cite{zhang2016learning} &55.40 &-&-&35.68\\
		Spindle \cite{zhao2017spindle} & 76.90& - &88.50& -\\
		P2S \cite{zhou2017point} & 70.72& 77.34 & -&44.27\\
		PrtAlign \cite{zhao2017deeply} &81.00 &88.50& 81.60 &63.40\\
		PDC \cite{su2017pose} &84.14 &- &88.70&63.41\\
		SSM \cite{bai2017scalable} &82.21& -& -&68.80\\
		JLML \cite{li2017person} &85.10&87.00&83.20&65.50 \\
		TriNet \cite{hermans2017defense} &84.92&-&-&69.14\\
		\hline
		\textbf{Ours}&\textbf{86.67} &  \textbf{89.90} &  \textbf{88.80}&\textbf{75.32}\\ 
		\hline
	\end{tabular} 
\end{table}
\begin{figure}[h!]
	\centering
	\includegraphics[width=0.99\linewidth]{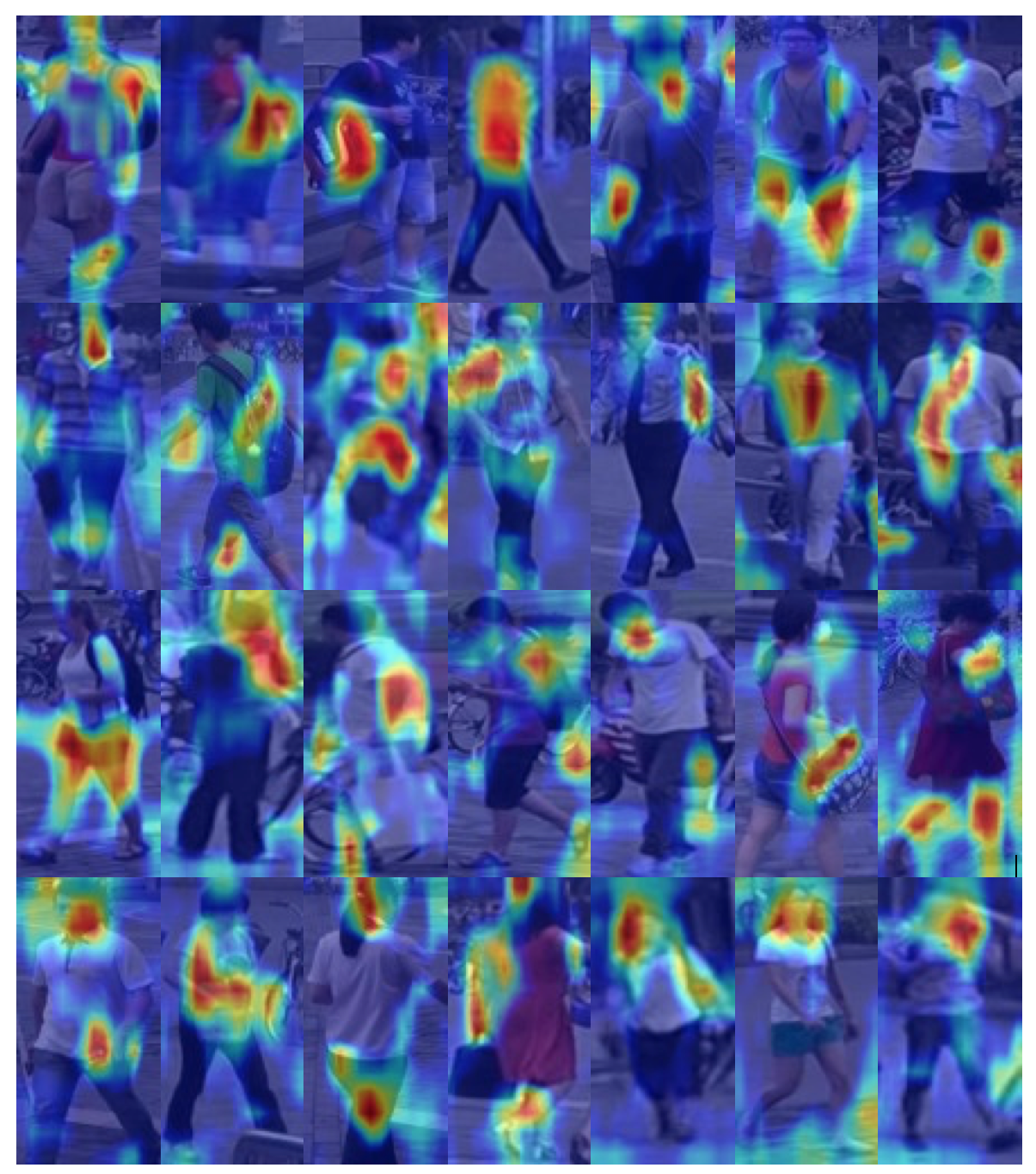}
	\caption{Visualization of the attention map produced by our proposed method}
	\label{fig:fig20}
\end{figure}

\begin{figure}[h!]
	\centering
	\includegraphics[width=1\linewidth]{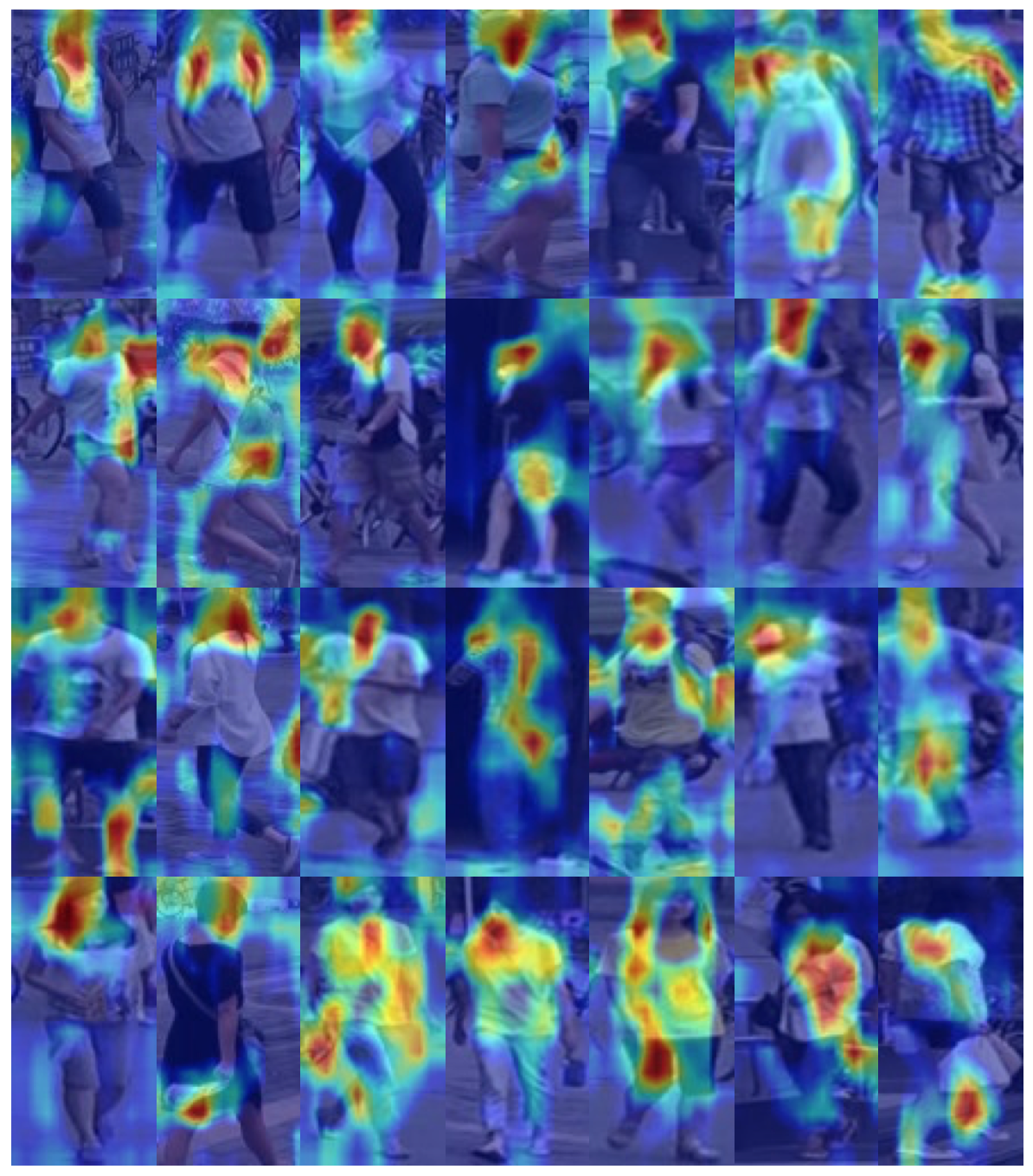}
	\caption{Visualization of the attention map produced by our proposed method (2)}
	\label{fig:fig35}
\end{figure}

\subsection{Interpretable Deep Retrieval Model}

The visualization of the attention map in our proposed Global attention net is shown in Figure \ref{fig:fig20} and \ref{fig:fig35}. These samples are part of the test query samples in Market 1501 dataset that are correctly re-identified by our model. 
These results show how the network is making its decisions and it thus makes our deep learning model more interpretable. For example, the visualization of the results shows how the attention model is able to focus on very detailed and discriminative parts of the input image (e.g., person's face, backpack, shoes, legs, t-shirts, things in their hands) Also, we can observe that by using our attention model, our re-identification system can successfully ignore the background clutter. 


\section{Conclusion}

In this paper, we introduced an attention mechanism for person re-identification task and we showed how paying attention to important parts of the person's image while still considering the whole image information, leads to highly discriminative feature embedding space and an accurate and interpretable person re-identification system. Furthermore, thanks to the computational efficiency resulting from the attention architecture, we would be able to use deeper neural networks and high resolution images in order to obtain higher accuracy.

\bibliographystyle{IEEEbib}
\bibliography{refs}

\end{document}